%% file: ms.tex
\documentclass[11pt,a4paper]{article}
\usepackage[table]{xcolor} 
\usepackage[hyperref]{acl2020}
\usepackage{times}
\usepackage{latexsym}
\usepackage{multirow,bigdelim,dcolumn,booktabs,array}
\usepackage{graphicx}
\usepackage[shortlabels]{enumitem}

\usepackage{todonotes}
\usepackage{booktabs}
\usepackage{dirtytalk}
\usepackage{microtype}
\usepackage{multirow}

\usepackage{amsmath}
\usepackage{amssymb}
\usepackage{subcaption}
\usepackage{xspace}

\usepackage{url}

\aclfinalcopy 
\def\aclpaperid{1553} 

\definecolor{very_likely}{HTML}{273253}
\definecolor{somewhat_likely}{HTML}{273253}
\definecolor{very_unlikely}{HTML}{c28004}
\definecolor{somewhat_unlikely}{HTML}{c28004}
\definecolor{background1}{HTML}{f7f7f7}

\title{Universal Decompositional Semantic Parsing}

\author{Elias Stengel-Eskin \\
  Johns Hopkins University  \\
\And
  Aaron Steven White  \\
  University of Rochester \\
\AND 
  Sheng Zhang \\
  Johns Hopkins University \\
\And
  Benjamin Van Durme \\
  Johns Hopkins University  \\ }

\date{}

\begin{document}
\maketitle
\begin{abstract}
 We introduce a transductive model for parsing into Universal Decompositional Semantics (UDS) representations, which jointly learns to map natural language utterances into UDS graph structures and annotate the graph with decompositional semantic attribute scores. We also introduce a strong pipeline model for parsing into the UDS graph structure, and show that our transductive parser performs comparably while additionally performing attribute prediction. By analyzing the attribute prediction errors, we find the model captures natural relationships between attribute groups.  
\end{abstract}

\section{Introduction}

A structured account of compositional meaning has been longstanding goal for both natural language understanding and computational semantics. To this end, a number of efforts have focused on encoding semantic relationships and attributes in a semantic graph---e.g. Abstract Meaning Representation \citep[AMR;][]{banarescu.l.2013}, Universal Conceptual Cognitive Annotation \citep[UCCA;][]{abend.o.2013}, and Semantic Dependency Parsing \citep[SDP;][]{oepen.s.2014, oepen.s.2015, oepen.a.2016}. 

\begin{figure*}[]
\includegraphics[width=\linewidth]{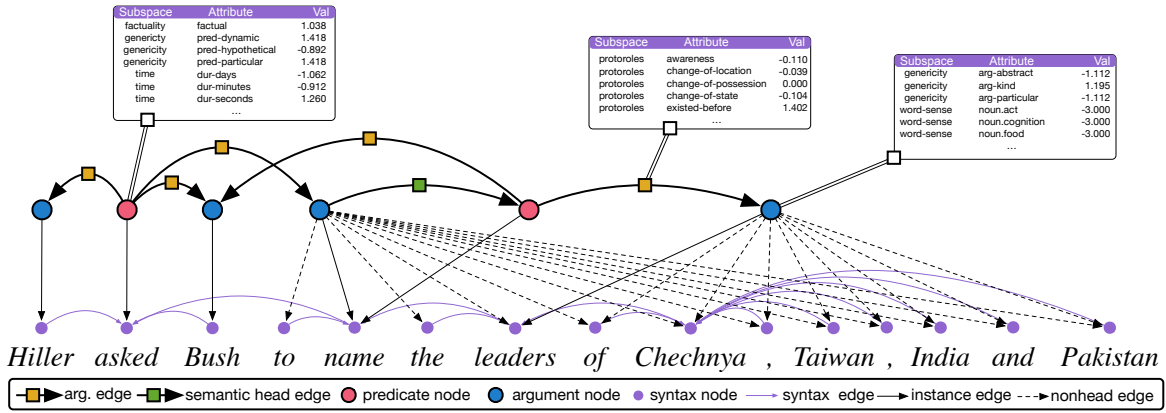}
\caption{The UDS graph structure. Semantic subgraph is outlined in black while the syntactic subgraph is annotated in pink. Node and edge attribute annotations are shown via annotations on argument and attribute edges.} 
\label{fig:decomp_graph}
\vspace{-1em}
\end{figure*}

In these formalisms, semantic information is typically encoded discretely, using nominal category labels for nodes and edges. This categorical encoding can make such formalisms brittle when presented with non-prototypical instances, and leads to challenges in coping with changing label ontologies and new datasets \citep{white.a.2019}. 
Furthermore, they are difficult to annotate, often requiring trained linguists and large annotation manuals. 

The Decompositional Semantics framework presents an alternative to categorical formalisms that encodes semantic information in a feature-based scheme---using continuous scales rather than categorical labels.
Starting with a feature-based semantic role representation rooted in \citealt{dowty.d.1991}'s (\citeyear{dowty.d.1991}) proto-role theory \citep{reisinger.d.2015, white.a.2016}, this framework has expanded to cover a wide variety of phenomena: event factuality \citep{rudinger.r.2018b}, genericity \citep{govindarajan.v.2019}, entity types \citep{white.a.2016}, and temporal relations \citep{vashishtha.s.2019}. 

While this rich array of annotation types has been separately modeled, no system yet exists for its joint prediction, which has only recently been made feasible by the introduction of Universal Decompositional Semantics v1.0 (UDS1.0). Presented by \citet{white.a.2019},
UDS1.0 normalizes all of these annotations, and incorporates them as node- and edge-level attributes in a single semantic graph whose structure is deterministically extracted from Universal Dependencies \citep[UD;][]{nivre.j.2015} syntactic parses via the PredPatt tool \citep{white.a.2016, zhang.s.2017}.\footnote{Available at \url{http://decomp.io}.} 
An example graph can be seen in Fig.~\ref{fig:decomp_graph}. 

We present the first joint UDS parser, which learns to extract both graph structures and attributes from natural language input. 
This parser is a sequence-to-graph transductive model which takes as input a sentence and outputs a UDS graph complete with node- and edge-level annotations. 

In contrast to the traditional semantic parsing paradigm, which shares its roots with syntactic parsing and rests on the assumption that the nodes in the graph correspond to tokens in the input---i.e. the graph is {\emph{lexicalized}}---the \textit{parsing-as-transduction} paradigm treats parsing as a sequence-to-graph problem. 
Rather than generating one sequence conditional on another sequence (sequence-to-sequence), we generate the nodes in a graph conditional on an input sequence, dynamically adding their edges during generation. 
As in sequence-to-sequence modeling, the supports of the input and output distributions---i.e. the input and output vocabularies---are not constrained to be identical. 

This has two benefits: first, post-hoc methods of obtaining alignments between input sequences and graphs---common especially in AMR parsing---are no longer required; and second, we are able to produce semantic graphs from arbitrary input vocabularies---allowing for future extensions to cross-lingual parsing \citep{zhang.s.2018a}. 
The parsing-as-transduction paradigm thus lends itself perfectly to UDS parsing, since the UDS protocol allows non-lexicalized (as well as cross-lingual) graphs, and these graphs may have nodes with multiple parents---i.e. re-entrant nodes---which pose problems for traditional tree-based methods but are handled natively by the transductive paradigm. 

We compare our end-to-end transductive parser against a strong pipeline system, finding that the parser slightly outperforms the pipeline while additionally learning to produce decompositional attribute scores. Our results are reflected in the UDS1.0 leaderboard at \url{http://decomp.io/leaderboards/}.

\section{Related Work}
\vspace{-0.5em}
\paragraph{Datasets}
\label{related}
\citet{reisinger.d.2015} introduce the Decompositional Semantics framework in the context of a corpus-based verification of \citeauthor{dowty.d.1991}'s seminal proto-role theory of semantic roles. This work was substantially expanded by \citet{white.a.2016}, who annotate for semantic proto-roles (SPR), word-sense, and temporal properties on top of semantic graphs extracted from English Web Treebank \citep[EWT;][]{bies.a.2012} UD parses using PredPatt \citep{white.a.2016,zhang.s.2017}.

\citeauthor{white.a.2016}'s EWT annotations are modeled by \citet{teichert.a.2017}, who present a CRF-based multi-label classifier for proto-role labelling, and \citet{rudinger.r.2018a}, who make use of an event-driven neural model. More recently, the annotation coverage for the same EWT data was expanded by \citet{vashishtha.s.2019} who annotate and model fine-grained temporal distinctions, and \citet{govindarajan.v.2019}, who add annotations and models for genericity---i.e. the degree of generality of events and entities in linguistic expressions. 

All of these efforts coalesce in \citet{white.a.2019}, which presents the first unified Decompositional Semantics-aligned dataset---Universal Decompositional Semantics v1.0 (UDS1.0)---containing all properties annotated on top of EWT parses with standardized train, validation, and testing splits and a native reader and query interface.

\paragraph{Parsing}
In most work on decompositional semantics, models are tasked with learning to predict attribute values, but not the structure of the graph. \citet{zhang.s.2018a} develop the first model for performing both graph parsing and UDS attribute prediction in a cross-lingual setting, where Chinese input sentences were transduced into UDS graphs derived from UD parses of the input's English translation.
This represents the first application of the parsing-as-transduction paradigm to a subset of UDS data as well as the introduction of a novel graph evaluation metric, $S$ which we describe in further detail in Section \ref{sec:s_metric}.
In contrast to the end-to-end approach presented here, \citeauthor{zhang.s.2018a} take a pipeline approach to parsing.

\citet{andreas.j.2013} recast semantic parsing in a tree formalism as a sequence-to-sequence problem.
Parsing-as-transduction, which extends this approach to directed acyclic graphs, has proven to be applicable in a variety of settings: \citet{zhang.s.2019a} use it to achieve state-of-the-art results in AMR parsing. 
These results are improved upon and shown to generalize to two other semantic formalisms (UCCA and SDP) by \citet{zhang.s.2019b}, which set new state-of-the-art benchmarks for AMR and UCCA. 
The former result was subsequently surpassed by \citet{cai.d.2020}, which applies a similar transductive approach, while the latter was surpassed by \citet{jiang.w.2019}.

Having both been subjects of SemEval tasks \citep{may.j.2016, may.j.2017, oepen.s.2019, hershcovich.d.2019}, there are a number of contrasting methods for both AMR and UCCA parsing. These include transition-based parsing system for AMR \citep{wang.y.2018, goodman.j.2016, damonte.m.2017, ballesteros.m.2017} and for UCCA \citep{hershcovich.d.2017}.
In a similar vein to \citet{zhang.s.2019b}, \citet{hershcovich.d.2018} convert multiple formalisms into a unified formalism and use multitask learning for improved UCCA parsing; however, the latter does so at a loss to performance on the other formalisms, while \citeauthor{zhang.s.2019b} achieve state-of-the-art results in AMR and UCCA simultaneously.
UCCA has also been shown to transfer to syntactic parsing: by converting UD parse trees into a format resembling UCCA, \citet{hershcovich.d.2018b} are able to apply a UCCA parser to both standard UD parses as well as enhanced UD parses, which contain re-entrant nodes. 

\section{Data}
 \begin{table*}
 \rowcolors{2}{gray!25}{background1}
 \begin{center}
  \begin{tabular}{ p{2cm}  p{6.25cm}  p{6.25cm} }
    \rowcolor{gray!50}
    {\textbf{Annotation}} & \textbf{Description} & \textbf{Examples} \\
    {\textbf{Factuality}} & Factuality inferences represent how likely (or unlikely) a listener thinks a scenario is to have occurred. &  {\textcolor{very_likely}{Jo \textbf{left} (3)}}, {\textcolor{very_unlikely}{Jo didn't \textbf{leave} (-3)}},
    {\textcolor{somewhat_unlikely}{Jo thought that Cole had \textbf{left} (-1)}} \\
    {\textbf{Genericity}} & Genericity refers to inferences about the generality of events or event participants. & \emph{Ex. property: genericity-pred-particular}: 
    \textcolor{very_likely}{Amy \textbf{ate} oats for breakfast today (3)}, \textcolor{very_unlikely}{Amy \textbf{ate} oats for breakfast every day (-3)} \\
    \textbf{Time} & Temporal inferences pertain to the duration of events. & \emph{Ex. property: time-dur-minutes}: \textcolor{very_unlikely}{Tom \textbf{left} (-3)}, \textcolor{very_likely}{Tom was \textbf{singing} (3)} \\
    \textbf{Word Sense} & UDS decomposes word sense, allowing multiple senses to apply to a given node. & \emph{Ex. property: supersense.person}: \textcolor{very_unlikely}{Sandy led \textbf{Rufus} by a leash (-3)}, \textcolor{very_likely}{\textbf{Sandy} led Rufus by a leash (3)} \\
    \textbf{Semantic Proto-Roles} & SPR properties are edge-level annotations that capture fine-grained semantic relations between predicates and arguments. & \emph{Ex. property: volition}:, \textcolor{very_unlikely}{\textbf{Derek broke} his arm (-3)}, \textcolor{very_likely}{\textbf{Derek broke} the wishbone (3)}  \\
    
  \end{tabular}
 \end{center}
 \vspace{-1em}
 \caption{Type descriptions and illustrative sentences for UDS properties predicted in this work. Example ratings in parentheses, bolding indicates the salient predicate/argument/edge. See \citet{white.a.2019} for further details.} 
  \label{tab:data}
  \vspace{-4mm}
\end{table*}
\label{data}

The UDS1.0 dataset is built on top of the UD-EWT data with three layers of annotations: UD parses, PredPatt graph structure, and decompositional semantic annotations on the edge and node level. 
In addition to specifying the syntactic head and head relation of every token in the input, UD parses include lexical features, such as word form, word lemma, and part-of-speech (POS) tag. This forms the \emph{syntactic graph}, which is lexicalized (each token is tied to a node in the graph). 
From these pieces of information, PredPatt outputs a set of predicates and their arguments. 

Each predicate and argument is tied via an {\textit{instance edge}} to a particular node in the syntactic graph. Because both predicates and arguments can consist of multi-word spans, there can be multiple instance edges leaving a semantic node.
The \emph{semantic graph} contains edges between predicates and arguments; in the case of clausal embedding, there can also be argument-argument edges. 
UDS1.0 includes ``performative'' speaker/author and addressee nodes, which model discourse properties of the sentence. These nodes are structural placeholders for future discourse-level annotations; as these properties have not yet been annotated, we have opted to remove them from the graphs.\footnote{Since these placeholder nodes are currently added deterministically, recovering them is also a deterministic operation.} 

The crowdsourced decompositional annotations tied to the semantic subgraph can be divided into node-level annotations and edge-level annotations. On the node level, annotations were collected for factuality, genericity, time, and entity type.
Edge-level annotations are in the space of semantic proto-roles, which are designed to provide a nuanced higher-dimensional substrate for notions of agency and patienthood. 
These are summarized in Table \ref{tab:data}, where purple indicates a high attribute score, while orange indicates a low score.
For further details on attribute types and data annotation, see \citet{white.a.2019} and the references therein.

\vspace{-0.5em}
\paragraph{Arborescence} Recall that the lowest level of the UDS graph (Fig.~\ref{fig:decomp_graph}) is a syntactic dependency parse. Modeling this level is out of scope for this work, as we are interesting in modeling the semantic structure and attributes. In order to train a parsing-as-transduction model, an arborescence---a hierarchical tree structure which has only edge and node annotations---is required. From the full UDS graph, we construct the arborescence by: 

\begin{figure}
\includegraphics[width=\columnwidth]{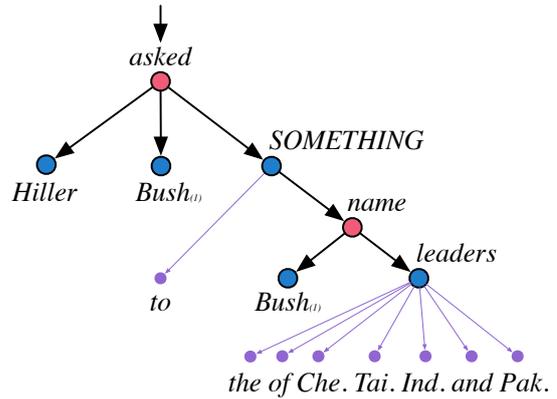}
\caption{Arborescence for graphs with object control.}
\label{fig:arborescence}
\vspace{-1em}
\end{figure}

\begin{enumerate}[(a),noitemsep,nolistsep]
    \item Assigning each semantic node a lexical label; this label is taken from the syntactic head that the semantic node dominates. The only exception to this is in the case of embedded clauses, where an argument node dominates an embedded predicate. Here, we follow PredPatt, assigning the label ``SOMETHING" to the embedded argument (c.f. Fig.~\ref{fig:arborescence}).
    \item Retaining all edges between semantic nodes as ``argument" edges, duplicating nodes in cases of re-entrancy (e.g. ``$\text{\emph{Bush}}_{\text{\emph{(1)}}}$'' in Fig.~\ref{fig:arborescence}). 
    \item Converting the deep syntactic structure into a shallow representation, where we introduce ``non-head" edges from the syntactic head (attached to a semantic node) to each node it dominates, and remove all other syntax-semantics edges. This effectively linearizes the yield of each semantic node (see Fig.~ \ref{fig:arborescence}).
\end{enumerate}

\section{Model}
Our model is based on the transductive broad-coverage parsing model presented in \citet{zhang.s.2019b}, which can be consulted for further details on the encoder, decoder, and pointer-generator modules.
The original parser is composed of six major modules: the encoder, the decoder embedding module, the target node module, the target label module, the head module, and the relation module. 
In this work we introduce two new modules: the node attribute module and the edge attribute module, as well a loss function for attributes. 

\vspace{-0.5em}
\paragraph{Encoder} The encoder module takes a concatenation of multiple input features: GloVe token embeddings \citep{pennington.j.2014}, POS tag embeddings, character CNN embeddings, and BERT \citep{devlin.j.2019} contextual embeddings (mean-pooled over subwords). These representations are passed through a stacked bidirectional LSTM  encoder, which has the following definition: 

\vspace{-6mm}
\begin{align*}
\mathbf{s}^l_t = \begin{bmatrix}
    \overrightarrow{\mathbf{s}}^l_t \\
    \overleftarrow{\mathbf{s}}^l_t
\end{bmatrix} = 
\begin{bmatrix}
\overrightarrow{\text{LSTM}}(\mathbf{s}_t^{l-1}, \mathbf{s}_{t-1}^t) \\
\overleftarrow{\text{LSTM}}(\mathbf{s}_t^{l-1}, \mathbf{s}_{t+1}^t)
\end{bmatrix}
\end{align*}

\noindent where arrows denote the LSTM direction, $t$ denotes the timestep, and $l$ denotes the layer of the stack. 

\vspace{-0.5em}
\paragraph{Decoder embedding module} In order to generate new semantic nodes and relationships, a method of embedding categorical semantic information is required. 
More formally, a semantic relation is given by a tuple $\langle u_i, d^u_i, r_i, v_i, d^v_i\rangle$, where $u_i$ denotes the ``head'' token of index $i$ and $v_i$ denotes the token at index $i$. 
Note that these tokens are the labels of nodes in the arborescence (see Fig~\ref{fig:arborescence}.)
$d_i^u$ and $d_i^v$ are the indices of $u_i$ and $v_i$, while $r_i$ is the relationship type between $v_i$ and $u_i$. 
The decoder embedding module embeds these categorical variables into real space, producing a tuple of vectors $\langle\mathbf{u}_i, \mathbf{d}_i^u, \mathbf{r}_i, \mathbf{v}_i, \mathbf{d}_i^v\rangle$. 
For node labels $\mathbf{u}_i$ and $\mathbf{v}_i$, we take the concatenation of GloVe and CharCNN features. 
$\mathbf{r}_i$, $\mathbf{d}_i^v$ and $\mathbf{d}_i^u$ are randomly initialized. 

\vspace{-0.5em}
\paragraph{Target Node Module} From the continuous embedding of a semantic relation $\langle\mathbf{u}_i, \mathbf{d}_i^u, \mathbf{r}_i, \mathbf{v}_i, \mathbf{d}_i^v\rangle $ we want to obtain a latent node representation $\mathbf{z}_i$. We initialize the hidden states of the $0^{\text{th}}$ layer and the hidden states of the $0^{\text{th}}$ state in each layer to
\begin{align*}
    \mathbf{h_i^0} &= [\mathbf{v}_i; \mathbf{d}^v_i]\\
    \mathbf{h_0^l} &= [\overleftarrow{\mathbf{s}}^l_1; \overrightarrow{\mathbf{s}}_n^l]
\end{align*}

\noindent respectively. Further, let $\mathbf{c}_i$ be a context vector over encoder states $\mathbf{s}_{1:n}^l$, defined as 
\begin{align*}
    \mathbf{a}_i^{\text{(enc)}} &= \text{softmax}\big(\text{MLP}^{\text{(enc)}} ([\mathbf{h}_i^l; \mathbf{s}_{1:n}^l]) \big)\\
    \mathbf{c}_i &= \mathbf{a}_i^T \mathbf{s}_{1:n}^l
\end{align*}
\newline

\noindent Let $\mathbf{h}_i^l$ and $\mathbf{z}_i$ be defined as follows:    \vspace{-2mm}
\begin{align*}
    \mathbf{z}_i &= \text{MLP}^{\text{(relation)}}([\mathbf{h}_i^l; \mathbf{c}_i; \mathbf{r}_i; \mathbf{u}_i;\mathbf{d}_i^u]) \\
    \mathbf{h}_i^l &= \text{LSTM}(\mathbf{h}_i^{l-1}, \mathbf{h}_{i-1}^l)
\end{align*}

\noindent where $\mathbf{z}_i$ can be thought as a representation of node $i$ in the graph, conditioned on previous nodes (via $\mathbf{h}_i^l$ as well as the input text via $\mathbf{c}_i$, the graph token (via $\mathbf{u}_i$ and  $\mathbf{d}_i^u$) and the relation type (via $r_i$). 

Using this representation $\mathbf{z}_i$, \citet{zhang.s.2019b} introduce an extended pointer-generator network \citep{see.a.2017} which computes the distribution over the next node label $v_{i+1}$: 
\begin{align*}
    [p_{\text{gen}}, p_{\text{enc}}, &p_{\text{dec}}] = \text{softmax}\big(\text{MLP}^{\text{(switch)}}(\mathbf{z}_i)\big) \\
    \mathbf{a}_i^{\text{dec}} &= \text{softmax}\big( \text{MLP}^{\text{dec}}([\mathbf{z}_{1:i}])\big)  \\
    \mathbf{\text{\bf{p}}}_i^{\text{(vocab)}} &= \text{softmax} \big( \text{MLP}^{(\text{vocab})} (\mathbf{z}_i)  \big) \\
    \label{eqn:final} \text{\bf{P}}(v_{i+1}) &=  p_{\text{gen}}\text{\bf{p}}_i^{\text{(vocab)}} \oplus p_{\text{enc}} \mathbf{a}^{\text{(enc)}}_i \oplus p_{\text{dec}}\mathbf{a}_i^{\text{(dec)}}
\end{align*}
\noindent From this last equation, we have that the generation of a new node is decomposed into three options: (1) generate a new node from a vocabulary of node labels, (2) copy a node label directly from the input sequence (lexicalization), or (3) copy a node label from a previously generated node (re-entrancy). 
\vspace{-0.5em}
\paragraph{Parsing modules} To obtain a parse from the node states $\mathbf{h}_{1:n}$, a head node and relation type must be assigned to each node $1:n$. In order to assign a head node, we instantiate two multi-layer perceptrons (MLPs): $\text{MLP}^{\text{(start)}}$ and $\text{MLP}^{\text{(end)}}$, where (start) denotes the starting node of the edge and (end) denotes its target. Using these MLPs, for node $i+1$ we obtain 
\begin{align*}
    \mathbf{h}_{i+1}^{\text{(start)}} &= \text{MLP}^{\text{(start)}} (\mathbf{h}_{i+1}^l) \\
    \mathbf{h}_{1:i}^{\text{(end)}} &= \text{MLP}^{\text{(end)}} (\mathbf{h}_{1:i}^l) \\
    \text{\bf{P}}(u_{i+1}) &= \text{softmax}\big( \text{BIAFFINE} (\mathbf{h}_{i+1}^{\text{(start)}}, \mathbf{h}_{1:i}^{\text{(end)}} )\big)
\end{align*}
\noindent The next relationship $r_{i+1}$ is computed in a similar fashion, also using two MLPs:
\begin{align*}
    \mathbf{h}_{i+1}^{\text{(rel-src)}} &= \text{MLP}^{\text{(rel-src)}} (\mathbf{h}_j^l) \\
    \mathbf{h}_{i+1}^{\text{(rel-tgt)}} &=  \text{MLP}^{\text{(rel-tgt)}} (\mathbf{h}_{i+1}^l) \\
    \text{\bf{P}}(r_{i+1}) &= \text{softmax}\big( \text{BILINEAR} ( \mathbf{h}_{i+1}^{\text{(rel-src)}}, \mathbf{h}_{i+1}^{\text{(rel-tgt)}}) \big)
\end{align*}

\noindent where $j$ is the index of the head assigned to the node indexed by $i+1$.\footnote{BIAFFINE is defined in \citet{dozat.t.2016}. $\text{BILINEAR}(x_1, x_2) = x_1 A x_2 + b$ where $A$ and $b$ are learned parameters.}

\vspace{-0.5em}
\paragraph{Node attribute module}
As noted in previous UDS projects, an important step in decompositional attribute annotation is determining whether a property applies in a given context. For example, factuality typically applies only to predicate nodes. Since all nodes (predicate and argument) are treated identically w.r.t. their semantic relations $\mathbf{z}_i$, this work introduces a two-fold node attribute model, which predicts whether a property $j$ applies to a node $i$ via a binary mask $\alpha_i^j$ as well as its value $\nu_i^j$. This module defines $\alpha_i^j$ and $\nu_i^j$ as follows:
\begin{align*}
    \text{\bf{P}}(\alpha_i^j) &= \text{sigmoid}\big( \text{MLP}^{\text{(node-mask)}} (\textbf{z}_i) \big) \\
    \nu_i^j &=  \text{MLP}^{\text{(node-attr)}} (\textbf{z}_i) 
\end{align*} 
\paragraph{Edge attribute module} As in the case of node attributes, edge attributes do not apply in all cases. Therefore, a similar bifurcation strategy is pursued with edge attribute prediction: we predict a binary attribute mask $\beta_{s,e}^j$ for attribute $j$ on edge $s\rightarrow e$ as well as an attribute value $\lambda_{s,e}^j$. These are given by:
\begin{align*}
    \mathbf{m}_{s,e}^{\text{(mask)}} &= \text{BILINEAR}^{\text{(mask)}}(\mathbf{h}_{s}^l, \mathbf{h}_{e}^l) \\
    \mathbf{m}_{s,e}^{\text{(attr)}}& = \text{BILINEAR}^{\text{(attr)}}(\mathbf{h}_{s}^l, \mathbf{h}_{e}^l) \\
    \text{\bf{P}}(\beta_{s,e}^j) &= \text{sigmoid}\big( \text{MLP}^{\text{(edge-mask)}} ( \mathbf{m}_{s,e}^{\text{(mask)}} ) \big) \\
    \lambda_{s,e}^j &=  \text{MLP}^{\text{(edge-attr)}} ( \mathbf{m}_{s,e}^{\text{(attr)}} ) 
\end{align*}
\vspace{-2.21em}
\paragraph{Training} The nodes in the graph are linearized in a pre-order traversal over the arborescence, which ensures that at prediction time, we have seen the potential antecendent of a node for target-side copying (e.g. $\text{\emph{Bush}}_{\text{\emph{(1)}}}$ in Fig.~\ref{fig:arborescence}), determining the order of semantic nodes in the graph. The syntactic children of these nodes are presented in the order they appear in the text. The loss functions for the node, head, and relation prediction modules are cross-entropy loss, while for the masks $\alpha$ and $\beta$ binary cross-entropy loss is used, since each position in the mask is a separate classification decision. The loss function used for $K$ attributes $\nu^{1:K}$ on $N$ nodes/edges is given by: 
\begin{align*}
    \tau(x) &= \begin{cases}
                0 &\text{ if } x \leq 0 \\
                1 &\text{ otherwise }\\
              \end{cases} \\
    \mathcal{L}_{\text{MSE}}(\nu, \nu^{*}) &= \frac{1}{NK}
    \sum\limits_{i=1}^N  \sum\limits_{j=1}^K c_i^j(\nu_i^j - \nu_i^{j*})^2& \\
\end{align*}
\begin{align*}
    \mathcal{L}_{\text{BCE}}(\nu, \nu^*) &=  \frac{1}{NK}
    \sum\limits_{i=1}^N  \sum\limits_{j=1}^K \bigg( \hspace{-0.3em}\tau(\nu^{j*}_i) \log(\tau(\nu^j_i))  \\
        &\hspace{1.5em}+ \big(1-\tau(\nu^{j*}_i)\big) \log \big(1 - \tau(\nu^j_i)\big) \bigg)& \\
    \mathcal{L}(\nu, \nu^*) &= \gamma \frac{2 * \mathcal{L}_{\text{MSE}}(\nu, \nu^*) * \mathcal{L}_{\text{BCE}}(\nu, \nu^*)}{\mathcal{L}_{\text{MSE}}(\nu, \nu^*) + \mathcal{L}_{\text{BCE}}(\nu, \nu^*)}
\end{align*}

\noindent where $\gamma$ is a scaling factor, $c_i^j$ is the annotator confidence for annotation $j$ on token $i$, $\nu$ is the set of predicted attributes, and $\nu^*$ is the set of true attributes. Note that inclusion of the confidence mask $c_i^j$ means the model only incurs loss on attributes annotated for a given node, since $c_i^j = 0$ when an annotation is missing (i.e. no MSE loss is incurred for attributes which do not apply to a node or edge); in the ``binary'' experimental setting, we replace $c_i^j$ with $\tau(c_i^j)$, removing the weighting but still masking out loss on un-annotated nodes. Also note than in the case of edges, the form of the loss is identical, but $\nu$ is replaced by $\lambda$, and $\alpha$ by $\beta$. 

This loss encourages the predicted attribute $\nu^j_i$ value to be close in value to the true value $\nu^{j*}_i$ via the mean-squared error criterion while concomitantly encouraging the predicted and reference values to share a sign via the thresholded cross-entropy criterion. Both node and edge attribute models are trained to predict attribute values independently, and that parameters are shared across attributes. This is central to our analysis in \S \ref{analysis}.

Following \citet{zhang.s.2019b} we train the structural parsing modules with coverage loss \citep{see.a.2017}. All models were trained to convergence using the Adam optimizer \citep{kingma.d.2014} with a learning rate of $0.001$. 

\section{Experiments}
\paragraph{Pipeline Model} 
Recall from Section \ref{data} that the semantic graph structure in UDS graphs is deterministically generated from PredPatt, which takes as input a UD parse and outputs a semantic graph structure. This leads to a strong pipeline model for the graph structure alone: running a high-performing UD parser---the Stanford UD parser \citep{chen.d.2014}---and passing its output through PredPatt to create a structure.\footnote{This structure is missing the core decompositional attributes but has both predicate and argument nodes. Additionally, the pipeline model fails to capture nominal heads of copular predicates (e.g. \textit{Jo is a \textbf{doctor}}), which are not returned by PredPatt but are added to the dataset as a preprocessing step in the genericity annotation task.} 
For this baseline, the only source of error is the UD parsing model, which for English performs very highly. 

\begin{table}[t!]
    \centering
    \begin{tabular}{l|ccc}
        \hline
         Method & P & R & F1 \\
        \hline
        Pipeline & 84.83 &75.22 & 79.74 \\
        Parser & 83.52 & 77.92 & 80.62 \\
        Parser (binary) & 84.97 & 78.52 & 81.62 \\ 
        \hline
    \end{tabular}
    \caption{Test set {\textit{S}} score precision, recall, and F1.} 
    \label{tab:structure_full}
    \vspace{-3mm}
\end{table}

\vspace{-0.5em}
\paragraph{\emph{S} Metric} \label{sec:s_metric} For evaluating the quality of output graph structures, Smatch \citep{cai.s.2013}, a hill-climbing approach to approximating the optimal matching between variables in two graphs, is commonly used. While Smatch can match categorial variables such as those found in meaning representations like AMR, it lacks a matching function for continuous variables such as decompositional attributes. To remedy this, \citet{zhang.s.2018a} introduce the $S$ metric, an extension to Smatch that allows for attribute matching. 

Using hill-climbing, we are able to match instance and attribute nodes and edges; instance nodes are matched via string match, while attribute similarity is given by $1 - \left(\frac{\nu_i - \nu_j}{\omega}\right)^2$
where $\omega = 6$ is the maximum possible difference between attributes, which are bounded on $[-3,3]$.\footnote{This function was found to produce more matches on UDS1.0 than the $e^{-\text{MAE}}$ function used by \citet{zhang.s.2018a}.}

\section{Results}

Table \ref{tab:attributes} shows the Pearson's correlation coefficient ($\rho$) and the F1 score computed on binarized responses for each node and edge attribute under the ``oracle'' decoding setting, where a gold graph structure is provided to the model. An asterisk denotes that $p < 0.05$, where $p$ is determined by a Student's t-test. 
F1 scores are obtained by binarizing continuous attribute predictions into positive and negative, following from the original UDS motivation found in \citet{dowty.d.1991}, where binary proto-role features were introduced. 
The binarization threshold was tuned per attribute on the validation set. 

The baseline column in Table \ref{tab:attributes} shows the binarized F1 score for the baseline attribute model, given by predicting the median attribute value for each attribute type at each position. Pearson's $\rho$ is undefined for this approach, as the variance of the predicted distribution is 0. The thresholds were similarly tuned on validation data for this baseline. 

Table \ref{tab:structure_full} shows \emph{S} metric (c.f. \S \ref{sec:s_metric}) precision, recall, and F1 score as computed on full arborescences with both semantics and syntax nodes. Our parser slightly outperforms the pipeline, with higher performance in the ``binary'' setting, where we exclude annotator confidence from the loss. 

Table \ref{tab:structure_semantics} shows precision, recall, and F1 score on semantics nodes alone. The first parser setting (syntax) reflects a parsing model trained on full graphs, and evaluated only on the semantic subgraphs of the produced graphs. The second parser (semantics) is directly trained on semantic subgraphs, with no syntactic nodes in the training graphs. The full parser performs comparably to the pipeline, while the parser trained specifically on semantics-only graphs outperforms the pipeline. However, the mean attribute $\rho$ of the syntactic parser ($0.3433$) exceeded that of the semantics-only parser ($0.3151$).

\begin{table}[h]
    \centering
    \begin{tabular}{l|ccc}
        \hline
         Method & P & R & F1 \\
        \hline
        Pipeline & 84.72  & 88.51  & 86.57  \\
        Parser (syntax) & 89.02 & 83.67 & 86.26 \\
        Parser (syntax, binary) & 89.74 & 86.00 & 87.83 \\
        Parser (semantics) & 91.28 & 87.23 & 89.21 \\
        Parser (sem., binary) & 91.10 & 84.59 & 87.73 \\
        \hline
    \end{tabular}
    \caption{Test set {\textit{S}} score precision, recall, and F1 on semantics nodes only, where (syntax) denotes a parser trained to predict full graphs (semantics nodes with non-head edges to syntax nodes) while (semantics) denotes model trained on semantics-only subgraphs. } 
    \label{tab:structure_semantics}
    \vspace{-3mm} 
\end{table}

\noindent Table \ref{tab:structure_attr} gives the \emph{S} metric results on full graphs predicted by the model, including attribute matching. 
The pipeline model is unable to perform this task because it predicts structure alone, without attributes. 
We see that training the parser with shared MLP and BILINEAR modules (i.e. $\text{MLP}^{\text{(mask)}} = \text{MLP}^{\text{(attr)}}$ and $\text{BILINEAR}^{\text{(mask)}} = \text{BILINEAR}^{\text{(attr)}}$) for both the attribute mask and attribute value heavily degrades the performance, while removing annotator confidence increases it slightly. 

\begin{table}[t]
    \centering
    \begin{tabular}{l|ccc}
        \hline
         Method & P & R & F1 \\
        \hline
        Shared & 79.52 & 32.48 & 46.12 \\
        Separate & 83.46 & 82.27 & 82.86 \\
        Separate (binary) & 84.19 & 84.19 & 84.19 \\
        \hline
    \end{tabular}
    \caption{Test set precision, recall, and F1 computed via {\textit{S}} score with attributes (syntactic nodes included)} 
    \label{tab:structure_attr}
    \vspace{-3mm} 
\end{table}

\input{new_f1_table.tex}

\section{Analysis} \label{analysis} Table \ref{tab:structure_full} suggests that the structural quality of the parses obtained by the parsing model presented here is slightly superior to that of pipeline model's parses, with Table \ref{tab:structure_semantics} indicating that the semantic component of the graph can be parsed significantly more accurately by our model. Taken together with Table \ref{tab:attributes}, we can conclude that the model is able to learn to jointly predict the graph structure and attributes. This is further reinforced by Table \ref{tab:structure_attr}. Note that the numbers reported in Tables \ref{tab:structure_full} and \ref{tab:structure_attr} are not directly comparable, as the scores in Table \ref{tab:structure_attr} incorporate the matching scores between attributes. 

Table \ref{tab:structure_semantics} shows that a parser trained on semantic subgraphs better recovers the subgraphs than a parser trained on full graphs whose outputs are postprocessed to remove syntactic nodes. However, the fact that the parser trained on full graphs achieves a higher Pearson's $\rho$ score indicates that the inclusion of syntactic nodes may provide additional information for predicting UDS attributes. 

In examining instances with an {\textit{S}} score below 50, we observe two trends: the input sentences are often ungrammatical, and for 63.82\% (on the validation set) the model predicts no output nodes. 

While the pipeline system does well on modeling semantic graph structure, it is by its definition unable to perform attribute parsing. In contrast, the results presented in Tables~\ref{tab:structure_attr} and \ref{tab:attributes} show that the parser can jointly learn to produce semantic graphs and annotate them with attributes. 

Finally, we find that while weighting the loss with the confidence scores has a small benefit in the semantics-only setting, it hurts overall attribute and structure prediction performance. This may be due to the relatively small size of the UDS dataset, which makes a strategy that is effectively weakening the loss signal at training time less effective.\footnote{All confidences are on $[0,1]$}

\begin{figure}[!h]
    \begin{subfigure}{\columnwidth}
    \centering
    \includegraphics[width=\columnwidth]{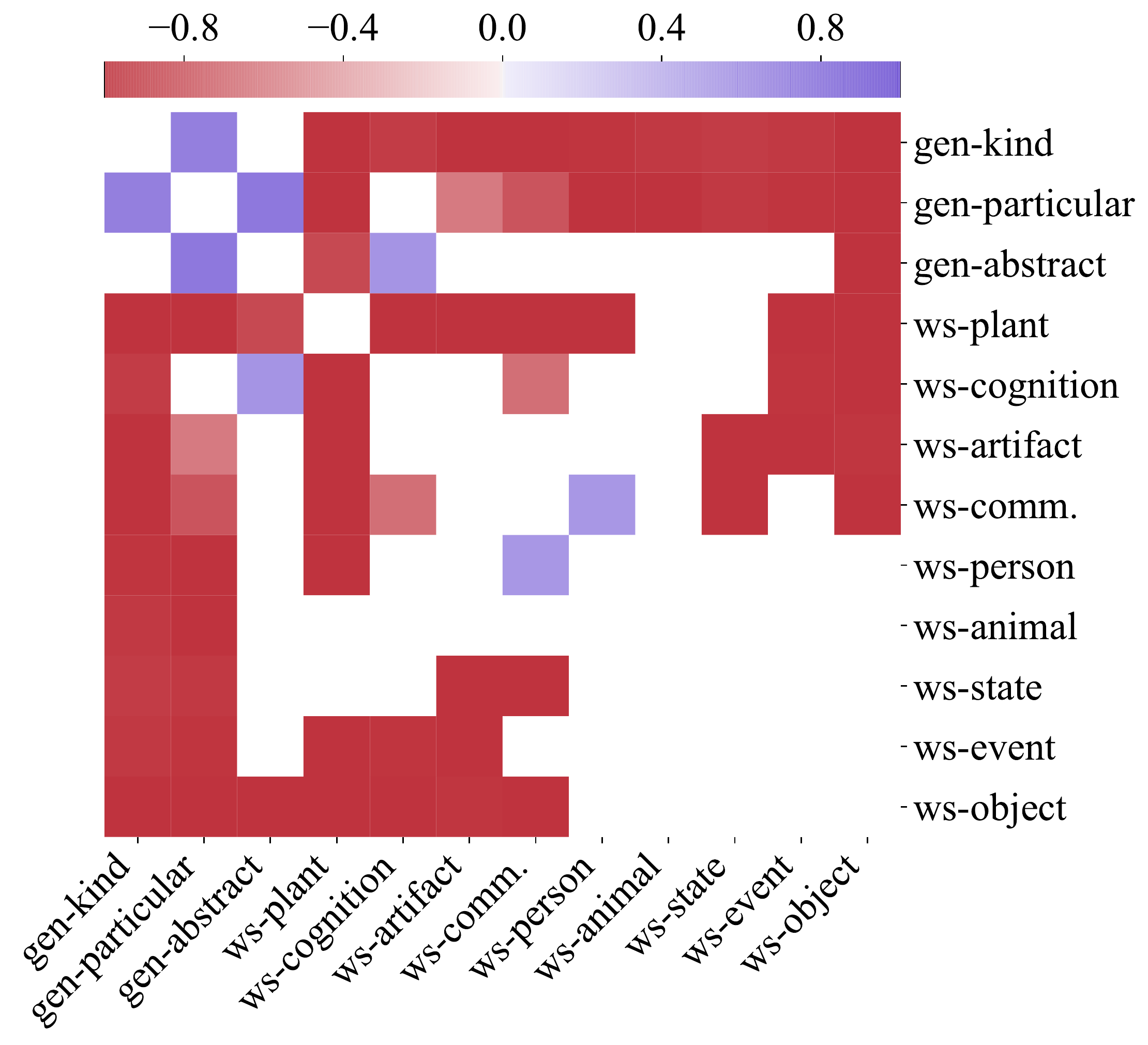}
    \caption{$\psi$ between argument-node attribute pairs. Subset of wordsenses used for readability.}
    \label{fig:node_arg}
     \end{subfigure}
\begin{subfigure}{\columnwidth}
    \centering
    \includegraphics[width=\columnwidth]{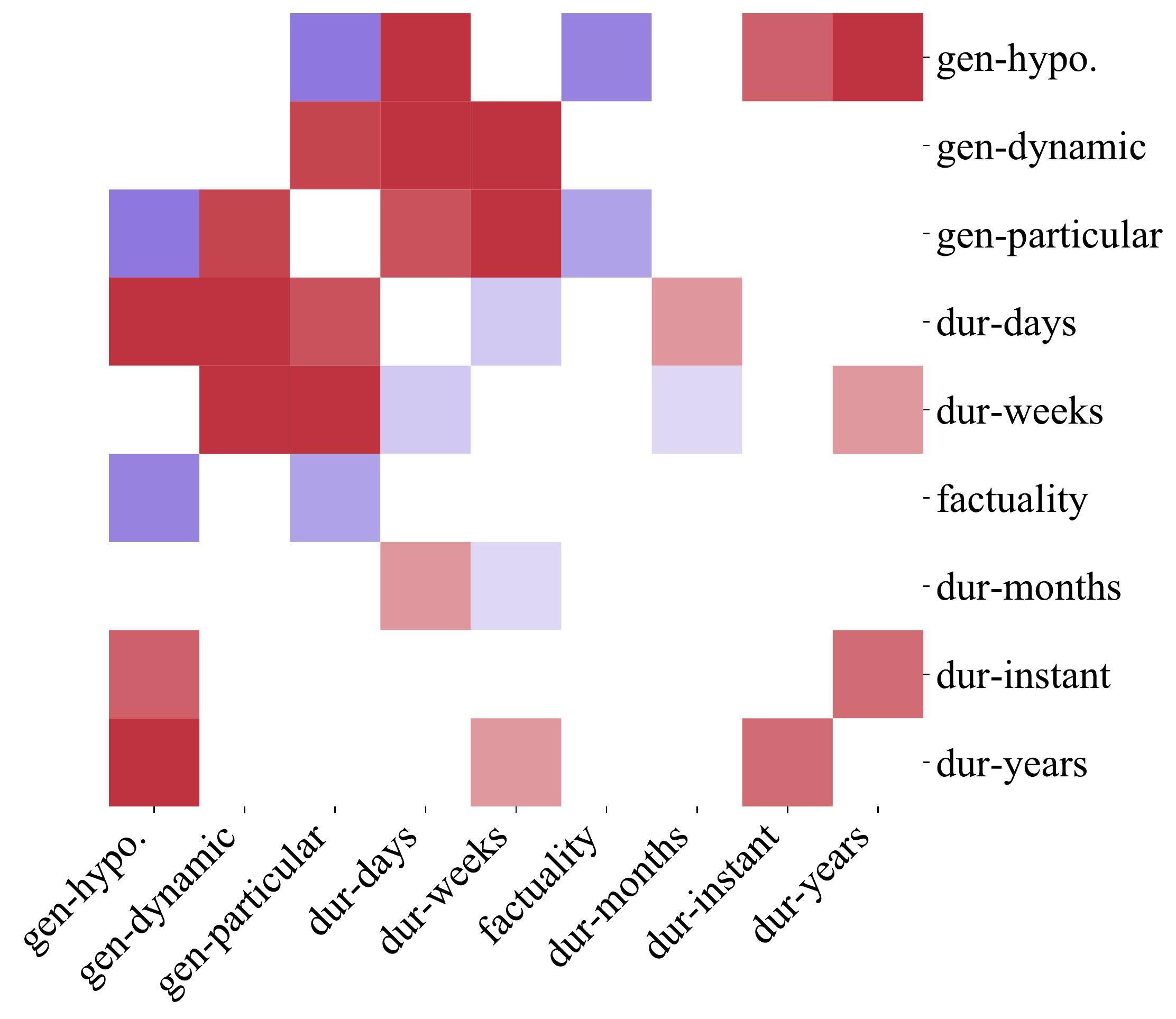}
    \caption{$\psi$ between predicate-node attribute pairs. Subset of time attributes used for readability. }
    \label{fig:node_pred}
\end{subfigure}
\begin{subfigure}{\columnwidth}
    \centering
    \includegraphics[width=\columnwidth]{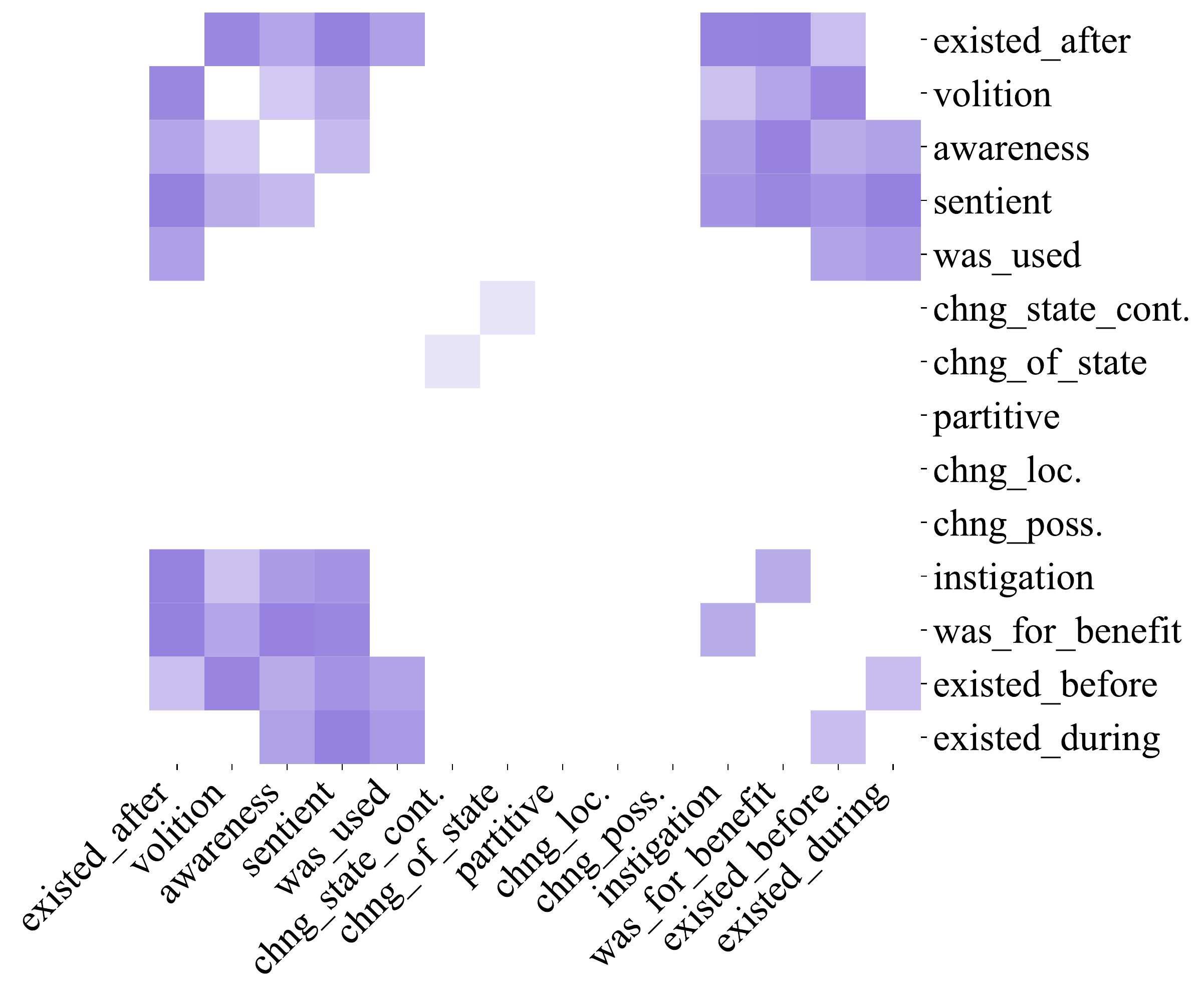}
    \caption{$\psi$ between all edge attribute pairs.}
    \label{fig:edge_pred}
\end{subfigure}
\caption{$\psi$ heatmaps for UDS1.0 attribute pairs}\label{fig:heatmaps}
\vspace{-8mm}
\end{figure}

Figs.~\ref{fig:node_arg}-\ref{fig:edge_pred} show the correlational strength coefficient between the true and predicted attributes under a forced decode of the graph structure. 
It is defined over property types indices $j$, $k$ with predicted attribute values $\nu_i^j$ and true values $\nu_i^{j*}$ as:

\vspace{-6mm}
\begin{align*}
\psi(j, k) &= \text{tanh}\big( 1 - \frac{|\text{corr}(\nu^j - \nu^{j*}, \nu^k - \nu^{k*}) | }{|\text{corr}(\nu^{j*}, \nu^{k*})|} \big)
\end{align*}
\vspace{-4mm}

\noindent where $\text{corr}(\nu^{j*}, \nu^{k*})$ is Pearson's correlation coefficient. Further details are given in Appendix~\ref{append:cov}. 

\begin{table*}[t]\centering
\begin{tabular}{l|r|c|c|c|c|c|c}
\hline
     Sentences & Property & (A) & Ours (A) & (B) & Ours (B) & (C) & Ours (C)  \\
\hline
    {\small{(A) \textbf{She} was untrained and,}}  & awareness & 3 & 3.04 & 1 & 3.69 & 5 & 3.68 \\
    {\small{in one botched job, \textit{killed} a client.}} & volition & 2 & 2.92 & 1 & 3.45 & 5 & 3.44 \\
    {\small{(B) \textbf{The antibody} then \textit{kills} the cell.}} & instigation & 5 & 3.08 & 5 & 3.39 & 5 & 3.37  \\ 
    {\small{(C) \textbf{An assassin in Colombia} \textit{killed} }} & sentience & 5 & 2.99 & 1 & 3.71 & 5 & 3.70 \\ 
    {\small{a federal judge on a Medellin street.}} & existed-after & 5 & 3.57 & 2 & 3.79 &  5 & 3.78 \\
\hline
\end{tabular}
\caption{Comparison of  gold properties from \citet{reisinger.d.2015} (on an ordinal scale from 1 to 5, with 3 as ``neutral'') and predicted properties (mapped to $[1,5]$) for sentences involving the predicate ``kills''. } 
\label{tab:qual}
\vspace{-5mm}
\end{table*}

 $\psi(i,j)$ reflects how well the model captures the strength of the correlations (either positive or negative) between two attribute types in the dataset: a positive value indicates that the model captures the correlation to some extent, with values closer to 1 implying better performance; a value of 0 indicates that the model does not capture the correlation at all, or that no significant interaction was present;
 a negative value indicates that the model makes systematic mistakes while predicting the two variables, e.g. when the model under-predicts the value of property $i$, it also under-predicts property $j$'s value. 
 A Bonferroni-corrected non-parametric bootstrap test ($1000$ replicants) was used for significance testing, with failing pairs being said to not be reliably different from 0 correlation.

Fig.~\ref{fig:node_arg} shows the model typically systematically under- or over-predicts the values for pairs of argument-node attributes, with most $\psi$ values close to -1. However, we do see positive correlations between some of the genericity annotations, as well as between {\textit{genericity-arg-abstract}}, which rates how conceptually abstract an argument is, and the {\textit{cognition}} wordsense, which applies to  abstract terms such as ``doubts'' and ``thoughts''.

In Fig.~\ref{fig:node_pred}, we again observe several negative $\psi$ values; however, some positive correlations can be seen between certain time properties, such as {\textit{duration-days}}, {\textit{duration-weeks}}, and {\textit{duration-months}}, as well as more strongly positive $\psi$'s between certain genericity annotations. 
The positive $\psi$ between {\textit{factuality}} and {\textit{genericity-hypothetical}} indicates the model has captured the commonalities between predicates with these annotations. 

In contrast to the node attributes, Fig.~\ref{fig:edge_pred} shows stronger results for edge attribute prediction, with all significant $\psi$'s being positive, and related attributes falling into clusters (e.g. {\textit{volition}}, {\textit{awareness}}, {\textit{sentience}}, or the {\textit{existed}} attributes) 

\paragraph{Qualitative examples}
Table~\ref{tab:qual} lists three sentences from \citet{reisinger.d.2015} along with a relevant subset of their  original SPR properties and values; the scale in \citeauthor{reisinger.d.2015} was ordinal from 1-5, with 1 corresponding to ``very unlikely,'' 5 to ``very likely,'' and 3 to ``neutral.'' 
Our model's predictions for the same sentences and properties are given as well, mapped onto $[1, 5]$. 
We first note that the structural component of the model is sufficiently strong that the correct predicate-argument edges were extracted during parsing, allowing for a direct comparison between the annotations by \citeauthor{reisinger.d.2015} and the parser's predictions. 
We see that while for sentence (C), the model captures at least the correct direction of the protorole annotations, it overgeneralizes these results to (B), where a more nuanced analysis is required. 
For (A), we see that on most attributes the model captures the desired binary direction of the inferences, but that it fails on \textit{sentience}. Overall, the model's predictions are weaker than the desired output, even when the  prediction is on the correct side of the midpoint, 3.
This might help explain the disparity between Pearson and F1 scores in Table~\ref{tab:attributes}, and represents a direction for future work. 
Note that to obtain attributes for (A) and (B), the threshold for the masks $\beta$ was dropped; ideally, this would not be required.  

\section{Conclusion}
The scalar valued, multi-attribute nature of UDS provides for a distinct structured prediction problem as compared to other existing representations.  
We have demonstrated how a transductive parsing paradigm that has achieved state-of-the-art results on other representations can be adapted to UDS1.0 structures and attributes, and have provided procedures for analysis, with the fine-grained nature of UDS allowing for investigating novel correlations and aspects of meaning. 
While UDS structures and various attribute types have been modeled separately  \citep{vashishtha.s.2019, govindarajan.v.2019, white.a.2016, rudinger.r.2018a, rudinger.r.2018b, zhang.s.2018a}, this work represents the first time all of these attributes and structures have been modeled jointly, and establishes a baseline for future efforts on UDS1.0.

We envision future efforts exploring the interactions between improving the underlying graph-structure prediction and ever-better correlations to human judgements on individual properties.

\section{Acknowledgements}
This work was supported by NSF Awards \#1749025 and \#1763705, DARPA LORELEI and AIDA, and IARPA BETTER. We thank the anonymous reviewers for their constructive feedback. The views and conclusions expressed herein are those of the authors and should not be interpreted as representing official policies or endorsements of DARPA, IARPA, or the U.S. Government.

\bibliography{decomp}
\bibliographystyle{acl_natbib}
\appendix
\include{appendix}

\end{document}

%% file: new_f1_table.tex
\begin{table}[!h]
    \centering
    \resizebox{\columnwidth}{!}{%
    \begin{tabular}{lll|c|c|c}
    \hline
     \multirow{2}{*}{} & \multirow{2}{*}{} & \multirow{2}{*}{Property} & Pearson's $\rho$ & F1  & F1 \\
    & &  &  (model) & (baseline) & (model)   \\ 
    \hline

\ldelim\{{44}{3mm}[\parbox{3mm}{\rotatebox[origin=c]{90}{node-level}}] &  & factuality-factual & 0.6479* & 75.15 & 84.46\\
 & \ldelim\{{6}{3mm}[\parbox{3mm}{\rotatebox[origin=c]{90}{genericity}}] & arg-abstract & 0.3392* & 40.04 & 48.05\\
 &  & arg-kind & 0.2145* & 67.61 & 67.54\\
 &  & arg-particular & 0.3347* & 83.10 & 84.62\\
 &  & pred-dynamic & 0.2469* & 72.49 & 71.19\\
 &  & pred-hypothetical & 0.3442* & 44.16 & 50.21\\
 &  & pred-particular & 0.1887* & 77.47 & 78.16\\
 & \ldelim\{{11}{3mm}[\parbox{3mm}{\rotatebox[origin=c]{90}{time}}] & dur-centuries & 0.1336* & 10.14 & 12.30\\
 &  & dur-days & 0.1802* & 68.72 & 68.21\\
 &  & dur-decades & 0.2383* & 29.89 & 34.19\\
 &  & dur-forever & 0.2524* & 37.93 & 38.58\\
 &  & dur-hours & 0.2227* & 73.66 & 73.61\\
 &  & dur-instant & 0.1761* & 55.98 & 51.90\\
 &  & dur-minutes & 0.3409* & 86.28 & 87.05\\
 &  & dur-months & 0.3204* & 63.25 & 64.42\\
 &  & dur-seconds & 0.2751* & 65.33 & 64.75\\
 &  & dur-weeks & 0.2475* & 54.02 & 55.41\\
 &  & dur-years & 0.4239* & 65.03 & 66.19\\
 & \ldelim\{{26}{3mm}[\parbox{3mm}{\rotatebox[origin=c]{90}{wordsense}}] & supersense-noun.Tops & 0.4660* & 7.34 & 40.00\\
 &  & supersense-noun.act & 0.6007* & 27.37 & 56.39\\
 &  & supersense-noun.animal & 0.3773* & 5.60 & 25.64\\
 &  & supersense-noun.artifact & 0.5617* & 23.12 & 52.79\\
 &  & supersense-noun.attribute & 0.4505* & 10.81 & 29.27\\
 &  & supersense-noun.body & 0.4543* & 1.53 & 42.86\\
 &  & supersense-noun.cognition & 0.5692* & 21.17 & 50.56\\
 &  & supersense-noun.communication & 0.6182* & 30.60 & 62.12\\
 &  & supersense-noun.event & 0.4233* & 5.80 & 33.61\\
 &  & supersense-noun.feeling & 0.2404* & 2.74 & 5.45\\
 &  & supersense-noun.food & 0.6773* & 7.15 & 67.72\\
 &  & supersense-noun.group & 0.5650* & 15.57 & 55.22\\
 &  & supersense-noun.location & 0.5118* & 7.81 & 55.64\\
 &  & supersense-noun.motive & 0.3447* & 0.62 & 50.00\\
 &  & supersense-noun.object & 0.2276* & 2.04 & 19.05\\
 &  & supersense-noun.person & 0.6091* & 15.74 & 61.25\\
 &  & supersense-noun.phenomenon & 0.2955* & 2.04 & 8.85\\
 &  & supersense-noun.plant & 0.0358 & 0.21 & 13.33\\
 &  & supersense-noun.possession & 0.5247* & 6.67 & 47.62\\
 &  & supersense-noun.process & 0.1292* & 1.13 & 3.96\\
 &  & supersense-noun.quantity & 0.4403* & 4.92 & 36.11\\
 &  & supersense-noun.relation & 0.2089* & 2.34 & 11.94\\
 &  & supersense-noun.shape & 0.0659* & 0.31 & 1.55\\
 &  & supersense-noun.state & 0.4877* & 11.36 & 36.17\\
 &  & supersense-noun.substance & 0.2411* & 1.43 & 3.64\\
 &  & supersense-noun.time & 0.5175* & 10.99 & 51.43\\
\ldelim\{{14}{3mm}[\parbox{3mm}{\rotatebox[origin=c]{90}{edge-level}}] & \ldelim\{{14}{3mm}[\parbox{3mm}{\rotatebox[origin=c]{90}{protoroles}}] & awareness & 0.6715* & 68.20 & 81.99\\
 &  & change-of-location & 0.1061* & 38.98 & 36.90\\
 &  & change-of-possession & 0.0452 & 14.93 & 20.00\\
 &  & change-of-state & 0.0448 & 42.59 & 37.21\\
 &  & change-of-state-continuous & 0.0793 & 31.47 & 27.69\\
 &  & existed-after & 0.3910* & 93.33 & 95.58\\
 &  & existed-before & 0.4802* & 91.60 & 92.31\\
 &  & existed-during & 0.3247* & 98.31 & 98.61\\
 &  & instigation & 0.3820* & 74.48 & 76.77\\
 &  & partitive & 0.0213 & 31.91 & 34.64\\
 &  & sentient & 0.6494* & 64.67 & 82.81\\
 &  & volition & 0.5501* & 63.79 & 79.86\\
 &  & was-for-benefit & 0.2389* & 59.87 & 62.11\\
 &  & was-used & 0.1608* & 86.64 & 89.00\\
\hline
\midrule
 & & \textit{macro-average} & 0.3433 & 37.20 & 50.66 \\
\bottomrule
    \end{tabular}}
     \caption{Pearson's $\rho$, baseline F1, and model F1 for each UDS attribute given gold test-set graph structures.} 
    \label{tab:attributes}
    \vspace{-4mm}
\end{table}

%% file: appendix.tex
\section{Derivation of $\psi$}
\label{append:cov}
The metric used in visualizations Fig.~\ref{fig:node_arg}-\ref{fig:edge_pred} is given by:

\begin{align*}
\psi(j, k) &= \text{tanh}\big( 1 - \frac{|\text{corr}(\nu^j - \nu^{j*}, \nu^k - \nu^{k*}) | }{|\text{corr}(\nu^{j*}, \nu^{k*})|} \big)
\end{align*}

\noindent where $\text{corr}(\nu^j - \nu^{j*}, \nu^k - \nu^{k*})$ and $\text{corr}(\nu^{j*}, \nu^{k*})$ are defined as follows: 

\begin{align*}
  \overline{\nu}^j &= \frac{1}{N} \sum_{i=1}^N \nu_i^j \\
    O_j &= \frac{1}{N} \sum_{i=1}^N (\nu_i^j - \nu_i^{j*})^2 \\
    \text{corr}&(\nu^j - \nu^{j*}, \nu^k - \nu^{k*}) = \\ &\frac{\frac{1}{N} \sum_{i=1}^N \big( (\nu_i^{j*} - \nu_i^j) (\nu_i^{k*} - \nu_i^{k} ) \big)}{\sqrt{O_j}\sqrt{O_k}} \\
    E_{j} &= \frac{1}{N} \sum_{i=1}^N (\nu_i^{j*} - \overline{\nu}_i^{j*})^2 \\
    \text{corr}&(\nu^{j*}, \nu^{k*}) = \\ &\frac{\frac{1}{N} \sum_{i=1}^N \big( (\nu_i^{j*} - \overline{\nu}_i^{j*})(\nu_i^{k*} - \overline{\nu}_i^{k*}) \big)  }{\sqrt{E_j} \sqrt{E_k}}\\
\end{align*}

Note that by this definition, $\psi$ is effectively a ratio of Pearson correlations, where the denominator is exactly the Pearson correlation between $\nu^{j*}$ and $\nu^{k*}$.